\documentclass[10pt, conference]{IEEEtran}
\usepackage[left=0.625in,right=0.625in,top=0.75in,bottom=0.99in]{geometry}
\IEEEoverridecommandlockouts
\usepackage{cite}
\usepackage{amsmath,amssymb,amsfonts}
\usepackage{graphicx}
\usepackage{color}
\usepackage{subfig}
\usepackage{stackengine}
\usepackage{comment}
\usepackage[normalem]{ulem}
\usepackage{caption}
\usepackage{tabularx}
\usepackage{bm}
\usepackage{dsfont}
\usepackage{mathtools}
\usepackage{lettrine}

\usepackage[flushleft]{threeparttable} 
\usepackage{booktabs}

\usepackage{textcomp}
\usepackage{xcolor}

\usepackage[ruled]{algorithm2e}
\SetKwRepeat{Do}{do}{while}%

\SetCommentSty{mycommfont}
\usepackage{algpseudocode}

\def\BibTeX{{\rm B\kern-.05em{\sc i\kern-.025em b}\kern-.08em
    T\kern-.1667em\lower.7ex\hbox{E}\kern-.125emX}}
    
\usepackage{xspace}
\usepackage{ifthen}
\usepackage[inline]{enumitem}

\usepackage{algcompatible}

\newboolean{showcomments}
\setboolean{showcomments}{true}
\ifthenelse{\boolean{showcomments}}
{
}

\usepackage{subfig}
\usepackage{fancyhdr}

\begin{document}

\title{A Lightweight Digital-Twin-Based Framework for Edge-Assisted Vehicle Tracking and Collision Prediction}

\author{Murat Arda Onsu$^1$, Poonam Lohan$^1$, Burak Kantarci$^1$, Aisha Syed$^2$, Matthew Andrews$^2$, Sean Kennedy$^2$
\thanks{$^{1}$School of Electrical Engineering and Computer Science,
University of Ottawa, Ottawa, ON, Canada. Emails: \{monsu022, ppoonam, burak.kantarci\}@uottawa.ca.}
\thanks{$^2$Nokia Bell Labs, 600 March Road,
Kanata, ON K2K 2E6, Canada. \newline 
Emails: \{aisha.syed, matthew.andrews, sean.kennedy\}@nokia-bell-labs.com}
}



\maketitle
\begin{abstract}
Vehicle tracking, motion estimation, and collision prediction are fundamental components of traffic safety and management in Intelligent Transportation Systems (ITS). Many recent approaches rely on computationally intensive prediction models, which limits their practical deployment on resource-constrained edge devices. This paper presents a lightweight digital-twin-based framework for vehicle tracking and spatiotemporal collision prediction that relies solely on object detection, without requiring complex trajectory prediction networks. The framework is implemented and evaluated in Quanser Interactive Labs (QLabs), a high-fidelity digital twin of an urban traffic environment that enables controlled and repeatable scenario generation. A YOLO-based detector is deployed on simulated edge cameras to localize vehicles and extract frame-level centroid trajectories. Offline path maps are constructed from multiple traversals and indexed using K-D trees to support efficient online association between detected vehicles and road segments. During runtime, consistent vehicle identifiers are maintained, vehicle speed and direction are estimated from the temporal evolution of path indices, and future positions are predicted accordingly. Potential collisions are identified by analyzing both spatial proximity and temporal overlap of predicted future trajectories. Our experimental results across diverse simulated urban scenarios show that the proposed framework predicts approximately 88\% of collision events prior to occurrence while maintaining low computational overhead suitable for edge deployment. Rather than introducing a computationally intensive prediction model, this work introduces a lightweight digital-twin-based solution for vehicle tracking and collision prediction, tailored for real-time edge deployment in ITS.
\end{abstract}

\begin{IEEEkeywords}  Collision Prediction, Path Estimation, Vehicle Tracking, Digital-Twin, YOLOv11, K-D Tree, Object Detection, ITS, Video Analysis.
\end{IEEEkeywords}

\section{Introduction} \label{sec:1}

Vehicle tracking and collision prediction are key components of Intelligent Transportation Systems (ITS), supporting traffic safety and management. While recent machine learning advances enable automated perception and prediction, many existing solutions rely on computationally intensive models such as large language models (LLMs) and Vision Transformers, which are difficult to deploy on resource-constrained edge devices due to limited computational resources and high inference latency. Cloud-based deployment alleviates this limitation but introduces additional challenges related to communication latency and bandwidth consumption \cite{onsu2025leveraging, onsu2025semantic}.

Lightweight edge-based solutions provide an effective alternative by performing inference directly on surveillance cameras, reducing latency and preserving privacy. Object detection is widely adopted in this context, as it efficiently localizes traffic participants using pixel-level bounding box information \cite{2,3}. In particular, YOLO (You Only Look Once-based models offer a favorable balance between accuracy and computational efficiency. In this work, we employ the YOLOv11 detector \cite{O10} to extract vehicle positions from video streams, associate vehicles with their driving paths, and estimate their future motion.

Reliable evaluation of collision-prediction methods in real-world traffic is challenging due to safety concerns and limited controllability. To address these challenges, we leverage Quanser Interactive Labs (QLabs), a high-fidelity digital twin of an urban traffic environment that enables controlled, repeatable, and scalable data generation. The proposed framework consists of four main stages. First, YOLO detections across multiple simulated driving scenarios are used to generate pixel-wise path representations. Second, a K-D tree–based association mechanism \cite{ram2019revisiting} efficiently matches detected vehicle positions to candidate paths, significantly reducing inference time compared to linear search. Third, consistent vehicle identifiers are assigned and future directions are estimated based on historical path progression. Finally, collision probability is computed by jointly analyzing the spatial proximity and temporal alignment of predicted future vehicle trajectories.
The main contributions of this work are summarized as follows:
\begin{itemize}
\item We leverage a high-fidelity digital twin environment (QLabs) to develop and evaluate a lightweight traffic surveillance and collision-prediction framework.
\item We propose a YOLO-based framework for future path estimation and spatiotemporal collision prediction that achieves 88\% collision detection accuracy in simulated urban scenarios.
\item We provide a detailed and reproducible methodology for vehicle ID assignment, path association, and collision probability estimation using video data.
\end{itemize}
The remainder of the paper is organized as follows. Section~II reviews related work. Section~III presents the proposed methodology. Section~IV discusses experimental results, and Section~V concludes the paper.

\section{Related Works} \label{sec:2}

Prior studies on Intelligent Transportation Systems (ITS) extensively explore traffic monitoring, vehicle tracking, future motion estimation, and collision prevention. YOLO-based object detection has been widely adopted due to its efficiency and real-time performance. For instance, YOLO has been applied to traffic flow estimation through vehicle counting in real time \cite{1}, and integrated into ITS pipelines for detection, tracking, and counting tasks \cite{2}. Hybrid approaches combining YOLO with classical tracking methods have also been investigated; in \cite{3}, YOLO detections are coupled with Kalman filtering for highway traffic flow estimation, while \cite{4} employs object detection and centroid tracking to identify wrong-way vehicles.

Several works extend YOLO-based detection toward collision-related analysis. In \cite{5}, YOLO and DeepSORT are combined for vehicle identification, tracking, speed estimation, and distance-based collision assessment. A real-time traffic monitoring framework integrating YOLO detection with virtual detection zones and geometric calibration is proposed in \cite{6}. To improve future motion prediction, learning-based temporal models have been incorporated, such as YOLO combined with LSTM networks for trajectory and position prediction \cite{7}, \cite{9}. Similarly, \cite{8} proposes a forward collision prediction framework using tracking-by-detection with an IOU-based tracker and particle-filter-based motion modeling. While these approaches improve predictive capability, they typically rely on additional tracking or temporal prediction modules, increasing computational complexity.

Digital twin technologies have recently emerged as powerful tools for controlled and reproducible evaluation of intelligent transportation solutions. CarTwin \cite{10} introduces a high-fidelity digital twin of in-vehicle CAN networks for realistic testing of vehicle dynamics and communication behavior. In \cite{11}, a digital-twin-based scenario generation framework is proposed for SOTIF assessment, enabling systematic reconstruction of safety-critical driving scenarios. In contrast to these works, our study leverages a digital twin of an urban traffic environment and integrates a YOLO-based lightweight framework for vehicle tracking, future path estimation, and spatiotemporal collision prediction, with explicit emphasis on reproducibility and edge deployment.

\section{Methodology}

This research adopts an object-detection-based approach to vehicle tracking, future path estimation, and collision probability analysis in urban traffic scenarios. Data generation is performed using QLabs, a digital twin of an urban environment in which diverse traffic configurations and environmental parameters are systematically defined and executed. The methodology relies on a video-based dataset that preserves both spatial and temporal dynamics, enabling effective modeling of vehicle motion, interactions, and speed variations.

The proposed framework follows a four-stage pipeline, with each stage formally defined by a corresponding algorithm. In the first stage, the YOLOv11 model is applied to collected video sequences to extract vehicle centroids and construct pixel-wise representations of predefined driving paths. In the second stage, detected vehicles are efficiently associated with candidate paths using a K-D tree–based nearest-neighbor search to reduce inference time. The third stage performs consistent vehicle ID assignment and future path estimation based on historical path progression. Finally, in the fourth stage, collision probabilities are computed by analyzing the spatiotemporal relationships among predicted future trajectories. The notation used throughout the algorithms is summarized in Table~\ref{tab:notation_compact}.

\begin{table}[t]
\centering
\caption{Notation Table}
\label{tab:notation_compact}
\fontsize{7.5}{7.5}\selectfont
\begin{tabular}{|c|p{7.4cm}|}
\hline
\textbf{Symbol} & \textbf{Description} \\
\hline
$\mathcal{Q}$ & Set of scenarios used for offline path generation (Alg.~1)\\\hline
$\Phi(\cdot)$ & YOLO-based Object detection function for input frames (Alg.~1) \\\hline
$I_k$ & Video frame at index $k$ during path generation (Alg.~1)\\\hline
$\mathcal{Z}_p$ & Aggregated set of vehicle centroids collected for path $p$ (Alg.~1) \\\hline
$\mathcal{R}_p$ & Ordered pixel coordinates of path $p$ (Alg.~1, 2, 4) \\\hline
$N_r$ & Number of resampled points used to construct $\mathcal{R}_p$ (Alg.~1)\\\hline

$v$ & Vehicle (track) identifier (Alg.~3--5) \\\hline
$p$ & Path (lane/zone) identifier (Alg.~2, 4, 5) \\\hline
$D_{\mathrm{path}}$ & Spatial threshold for path association (Alg.~2)\\\hline
$\mathcal{F}$ & Set of path definition files (Alg.~1, 2) \\\hline
$\mathbf{r}_i^p$ & $i$-th pixel coordinate on path $p$ (Alg.~2, 4) \\\hline
$N_p$ & Number of pixels defining path $p$ (Alg.~2) \\\hline
$\mathcal{K}_p$ & KD-tree constructed from $\mathcal{R}_p$ (Alg.~2, 5) \\\hline
$\mathbf{c}$ & Query vehicle centroid (Alg.~2) \\\hline
$\mathbf{c}_i^t$ & Centroid of detection $i$ at frame $t$ (Alg.~3) \\\hline
$t$ & Discrete frame index (Alg.~3) \\\hline
$T_f$ & Total number of processed frames (Alg.~1,3) \\\hline
$\mathcal{T}_v$ & Observed centroid trajectory of vehicle $v$ (Alg.~3) \\\hline
$\mathcal{A}$ & Set of active vehicle tracks (Alg.~3) \\\hline
$\mathcal{U}$ & Set of inactive/recycled track identifiers (Alg.~3) \\\hline
$D_{\mathrm{trk}}$ & Distance threshold for vehicle tracking (Alg.~3) \\\hline
$\mathcal{H}_v$ & Path–index association history of vehicle $v$ (Alg.~4) \\\hline
$\widetilde{\mathcal{H}}_v$ & Downsampled association history (Alg.~4) \\\hline
$i_p^{(k)}$ & Associated index on path $p$ at history step $k$ (Alg.~4) \\\hline
$L,K$ & Temporal downsampling parameters (Alg.~4) \\\hline
$N$ & Prediction horizon (number of future points) (Alg.~4) \\\hline
$\widehat{\mathcal{Y}}_v$ & Predicted future paths for vehicle $v$ (Alg.~4, 5) \\\hline
$\widehat{\mathbf{R}}_{v,p}$ & Predicted trajectory of vehicle $v$ on path $p$ (Alg.~4, 5) \\\hline
$T$ & Total prediction duration (continuous time) (Alg.~5) \\\hline
$D$ & Spatial collision distance threshold (Alg.~5) \\\hline
$\Delta t$ & Temporal collision tolerance (Alg.~5) \\\hline
$N_{\mathrm{comb}}$ & Number of evaluated path combinations (Alg.~5) \\\hline
$N_{\mathrm{col}}$ & Number of colliding path combinations (Alg.~5) \\\hline
$\Pr_{\mathrm{col}}$ & Estimated collision probability (Alg.~5) \\\hline
$\mathcal{S}$ & Collision summary set (Alg.~5) \\\hline
\end{tabular}
\end{table}

\subsection{Digital Twin-based Data Collection} \label{sec: quanser}
\begin{figure}
    \centering
    \includegraphics[width=0.90\linewidth,height=6.0cm, keepaspectratio]{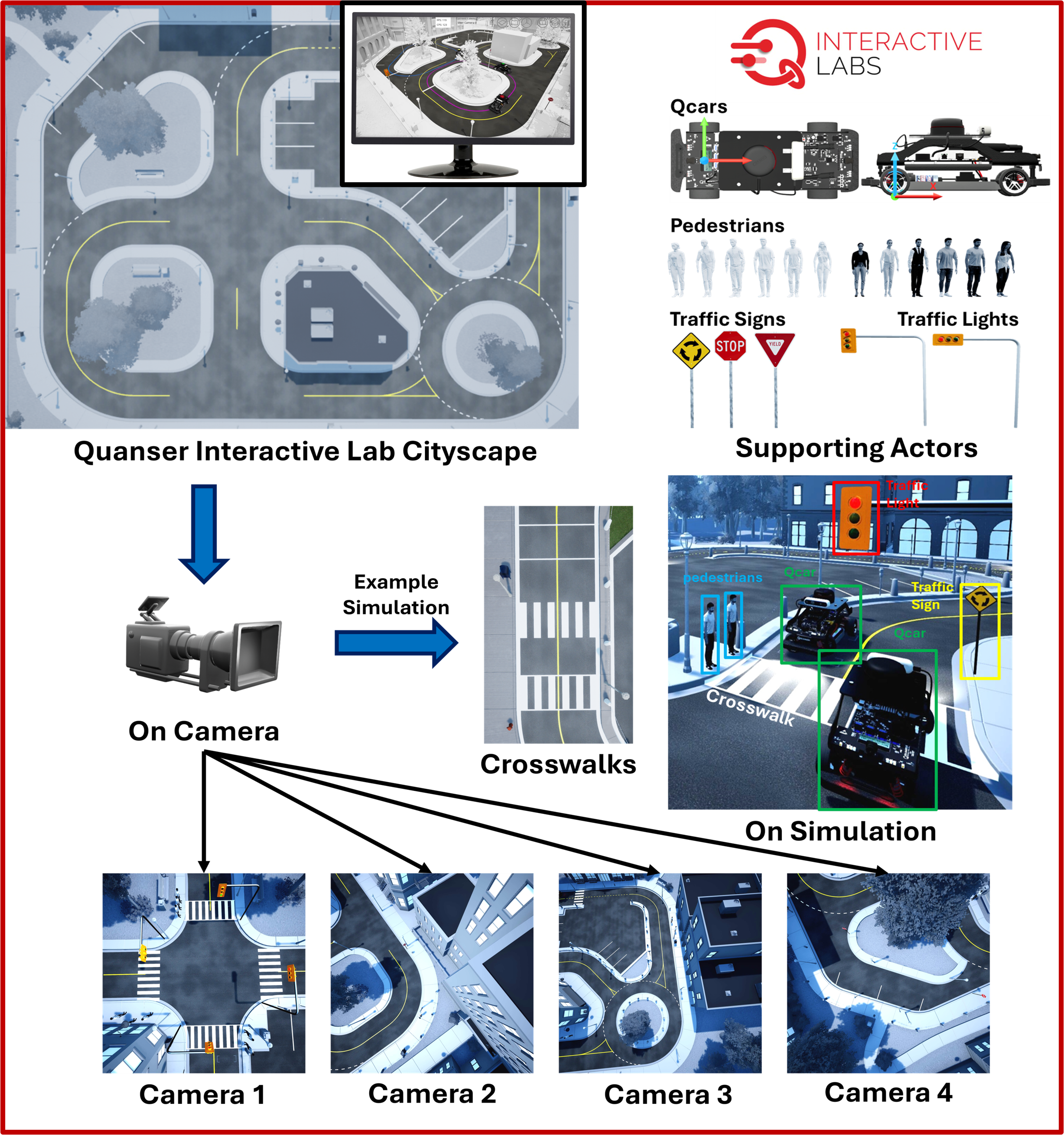}\vspace{0.1cm}
    \caption{Components and Architecture of digital Twin of Urban Traffic Environment: Quanser Interactive Labs (QLabs)}
    \label{fig:environment}
\end{figure}

The proposed framework requires diverse traffic scenarios involving multiple vehicles and interactions. Conducting such experiments in real-world environments, particularly repeated collision events, congestion, and multi-camera observations, is challenging due to safety risks and limited controllability. Therefore, data generation and evaluation are performed using QLabs, a high-fidelity digital twin of an urban traffic environment \cite{quanser_python_api_2024}. QLabs provides a controlled and configurable simulation platform with realistic traffic components, including roads, vehicles, pedestrians, traffic signs, traffic lights, and crosswalks, as illustrated in Fig.~\ref{fig:environment}.

Fig.~\ref{fig:environment} shows an overview of the QLabs environment, road network, and simulation components. The primary vehicle entity is the \textit{QCar}, which represents a real-world vehicle. During runtime, QCars follow predefined routes using an integrated driving model until route completion. The environment is initialized with an empty road layout and can be augmented with traffic signs (roundabout, stop, yield), two traffic light models, three crosswalk models, and six pedestrian models with corresponding low-resolution representations.
To obtain a comprehensive view of the traffic scene, four cameras are deployed to cover different regions of the environment. These cameras record video streams during runtime, and a video generation pipeline captures frames at 5~ms intervals to produce high-resolution ($2048 \times 2048$) video data. QLabs supports seamless integration with MATLAB/Simulink and Python, enabling efficient incorporation of the collected data into the proposed perception and analysis pipeline.

\subsection{YOLOv11 Model for Vehicle Detection}
The proposed framework relies on real-time object detection to localize vehicles in urban traffic scenes. In this work, vehicle detection is performed using the YOLO model, which is widely adopted in computer vision for real-time applications due to its high inference speed and single-pass detection strategy. YOLO is integrated into the simulated camera pipeline to accurately detect the positions of QCars in each video frame.

We employ the YOLOv11 model, as it provides improved detection performance while maintaining computational efficiency suitable for edge deployment \cite{O10}. The YOLOv11 architecture consists of three main components: a backbone, a neck, and a head. The backbone extracts hierarchical feature representations from input images using convolutional neural networks (CNNs). The neck aggregates multi-scale features to enhance robustness to object size variation, while the head performs bounding box regression and object classification to produce final detection outputs. During runtime, YOLOv11 processes each frame independently to produce vehicle bounding boxes, whose centroids are used as compact representations of vehicle positions. These detections serve as the input for subsequent vehicle tracking, path association, and motion estimation stages.

\subsection{Vehicle Path Map Generation} 
In this stage, vehicular paths and their pixel-wise representations are generated from YOLO detections across multiple scenarios. These path definitions are subsequently used for future direction estimation and collision probability analysis. The path generation procedure is summarized in Alg.~\ref{alg:path_generation}.  

\begin{algorithm}
\caption{\small{ Multi-Scenario Vehicle Path Generation}}
\label{alg:path_generation}
\fontsize{7.5}{7.7}\selectfont
\KwIn{$\mathcal{Q}$, $\mathcal{P}$, $\Phi(I_k)$, $T_f$, $N_r$}
\KwOut{$\mathcal{F}$, $\{\mathcal{R}_p\}_{p\in\mathcal{P}}$}

$\mathcal{F}\gets\emptyset$\;

\ForEach{$p\in\mathcal{P}$}{
    $\mathcal{Z}_p\gets\emptyset$\;
    \ForEach{$q\in\mathcal{Q}$}{
        Initialize traversal of route $p$ under scenario $q$\;
        \For{$k=1$ \KwTo $T_f$}{
            $\mathcal{B}_k\gets\Phi(I_k)$\;
            Select $\mathbf{b}_k\in\mathcal{B}_k$\;
            $\mathbf{c}_k\gets\big(\frac{x_{1,k}+x_{2,k}}{2},\frac{y_{1,k}+y_{2,k}}{2}\big)$\;
            $\mathcal{Z}_p\gets\mathcal{Z}_p\cup\{\mathbf{c}_k\}$\;
        }
    }
    $\mathcal{R}_p\gets\mathrm{Resample}(\mathcal{Z}_p,N_r)$\;
    Write $f_p$ with $\mathcal{R}_p$ as $(x,y)$ pairs\;
    $\mathcal{F}\gets\mathcal{F}\cup\{f_p\}$\;
}
\Return $\mathcal{F},\{\mathcal{R}_p\}$
\end{algorithm}

Alg. \ref{alg:path_generation} describes the offline construction of path definitions used by the subsequent tracking and prediction stages. For each predefined path $p\in\mathcal{P}$, multiple driving scenarios $q\in\mathcal{Q}$ are executed in which a vehicle is commanded to traverse the same route under varying conditions. During each traversal, image frames are processed sequentially, and a YOLO-based detector $\Phi(\cdot)$ is applied to each frame $I_k$ to extract bounding boxes corresponding to the traversing vehicle. The vehicle position at frame $k$ is approximated by the centroid of the detected bounding box, computed as $\mathbf{c}_k=\big(\frac{x_{1,k}+x_{2,k}}{2},\frac{y_{1,k}+y_{2,k}}{2}\big)$. 
All centroid samples collected across scenarios are aggregated into a raw spatial point set $\mathcal{Z}_p$, which captures the geometric variability of the path induced by different runtime. The aggregated point set is then resampled to obtain an ordered representation $\mathcal{R}_p$ at a desired spatial resolution. The resulting path representations are stored as path files $\mathcal{F}$ and are subsequently used for K-D tree construction and online path association.

\subsection{Vehicle Nearest Path Association Using K-D Tree}

Once all path definitions are constructed, each detected vehicle can be associated with a candidate path based on its current YOLO-derived centroid. The path association procedure is detailed in Alg.~\ref{alg:kdtree_assoc}. To enable efficient association, a K-D tree (K-Dimensional Tree), a space-partitioning data structure for nearest-neighbor and range queries, is built over the pixel coordinates defining each path. Given a query vehicle centroid, the K-D tree retrieves the nearest path point and its corresponding index in logarithmic time, avoiding exhaustive distance computations across all path pixels. Compared to linear search with $\mathcal{O}(N)$ complexity, the K-D tree reduces the average search cost to $\mathcal{O}(\log N)$, significantly accelerating path association and collision detection, particularly when operating on densely sampled spatial trajectories.

\begin{algorithm}
\caption{\small{K-D Tree-based Vehicle Path Association}}
\label{alg:kdtree_assoc}
\fontsize{7.5}{7.7}\selectfont
\KwIn{$\mathcal{F}$, $\mathbf{c}=(x,y)$, $D_{\mathrm{path}}$}
\KwOut{Associated path set $\mathcal{P}$}

$\mathcal{K} \gets \emptyset$\;

\ForEach{$f \in \mathcal{F}$}{
    Parse pixel coordinates $\mathcal{R}_p = \{\mathbf{r}_i^p\}_{i=1}^{N_p}$ from $f$\;
    \If{$N_p > 0$}{
        $\mathcal{K}_p \gets \mathrm{KDTree}(\mathcal{R}_p)$\;
        $\mathcal{K} \gets \mathcal{K} \cup \{(p,\mathcal{K}_p)\}$\;
    }
}

$\mathcal{P} \gets \emptyset$\;

\ForEach{$(p,\mathcal{K}_p) \in \mathcal{K}$}{
    $(d_p,i_p) \gets \arg\min_i \|\mathbf{c}-\mathbf{r}_i^p\|_2$\;
    \If{$d_p \le D_{\mathrm{path}}$}{
        $\mathcal{P} \gets \mathcal{P} \cup \{(p,i_p)\}$\;
    }
}

\Return $\mathcal{P}$
\end{algorithm}



Alg. \ref{alg:kdtree_assoc} performs vehicle-path association by querying a K-D tree representation of each path using the vehicle centroid $\mathbf{c}=(x,y)$. First, it initializes an empty collection $\mathcal{K}$ to store the K-D trees. For each path file $f\in\mathcal{F}$, the algorithm parses the ordered pixel coordinates $\mathcal{R}_p=\{\mathbf{r}_i^p\}_{i=1}^{N_p}$. If the path is non-empty ($N_p>0$), it constructs a KD-tree $\mathcal{K}_p\gets\mathrm{KDTree}(\mathcal{R}_p)$ and inserts the pair $(p,\mathcal{K}_p)$ into $\mathcal{K}$.
Next, the algorithm initializes the association set $\mathcal{P}\gets\emptyset$. For each stored pair $(p,\mathcal{K}_p)\in\mathcal{K}$, it queries $\mathcal{K}_p$ with the centroid $\mathbf{c}$ to obtain the nearest path point and its index, equivalently expressed as $(d_p,i_p)\gets\arg\min_i\|\mathbf{c}-\mathbf{r}_i^p\|_2$. The candidate path $p$ is accepted if the nearest distance satisfies $d_p\le D_{\mathrm{path}}$, in which case the association tuple $(p,i_p)$ is appended to $\mathcal{P}$. Finally, the algorithm returns $\mathcal{P}$, which may contain one or multiple paths depending on the spatial proximity of $\mathbf{c}$ to the available path geometries.

\subsection{Vehicle Tracking and Future Path Estimation}
 In this part, we generate two essential algorithm: one for consistent vehicle ID Assignment for tracking and the other one is estimating future path or direction. ID assignment algorithm can be seen in Alg. \ref{alg:vehicle_tracking}. 

\begin{algorithm}
\caption{\small{Vehicular Tracking and ID Assignments}}
\label{alg:vehicle_tracking}
\fontsize{7.5}{7.7}\selectfont
\KwIn{$\{\mathbf{c}_i^t\}_{t=1}^{T_f}$, $D_{\mathrm{trk}}$}
\KwOut{Tracks $\{\mathcal{T}_v\}$}

$\mathcal{A}\gets\emptyset,\;\mathcal{T}\gets\emptyset,\;\mathcal{U}\gets\emptyset$\;

\For{$t=1$ \KwTo $T_f$}{
    $\mathcal{C}_t\gets\{\mathbf{c}_i^t\}$, \;$\mathbf{c}_i^t=\big(\frac{x_{1,i}+x_{2,i}}{2},\frac{y_{1,i}+y_{2,i}}{2}\big)$\;
    Mark all $v\in\mathcal{A}$ unmatched\;
    \ForEach{$\mathbf{c}_i^t$}{
        $(v^\star,d^\star)\gets\arg\min_{(v,\mathbf{c}_v)\in\mathcal{A}}\|\mathbf{c}_i^t-\mathbf{c}_v\|_2$\;
        \eIf{$d^\star\le D_{\mathrm{trk}}$}{
            $\mathbf{c}_{v^\star}\gets\mathbf{c}_i^t,\;\mathcal{T}_{v^\star}\cup\{\mathbf{c}_i^t\}$\;
        }{
            $v_{\mathrm{new}}\gets(\mathcal{U}\neq\emptyset)?\mathcal{U}:\text{new}$\;
            $\mathcal{T}_{v_{\mathrm{new}}}\gets[\mathbf{c}_i^t],\;\mathcal{A}\cup\{(v_{\mathrm{new}},\mathbf{c}_i^t)\}$\;
        }
    }
    Remove unmatched $v$ from $\mathcal{A}$ and add to $\mathcal{U}$\;
}
\Return $\{\mathcal{T}_v\}$
\end{algorithm}

Consistent ID assignment is the method of putting an ID to the each vehicle which doesn't change during the runtime. It allows us to individually track each vehicle and store their informations separately. Alg. \ref{alg:vehicle_tracking} assigns persistent vehicle identifiers $v$ to frame-wise centroid detections $\{\mathbf{c}_i^t\}_{t=1}^{T_f}$ based on spatial proximity. This tracking stage is required to convert independent per-frame detections into temporally consistent trajectories $\{\mathcal{T}_v\}$, which are subsequently used for path association history construction and future motion analysis.
At each frame $t$, detected centroids $\mathbf{c}_i^t$ are matched to the set of active tracks $\mathcal{A}$ by minimizing the Euclidean distance to the last known centroid of each track. A detection is associated with an existing track if the distance is below the threshold $D_{\mathrm{trk}}$; otherwise, a new track identifier is assigned (optionally reusing an identifier from the inactive set $\mathcal{U}$). Tracks in $\mathcal{A}$ that remain unmatched are deactivated and moved to $\mathcal{U}$. The output is a collection of centroid trajectories $\{\mathcal{T}_v\}$, each representing the observed motion of a single vehicle over time.
In each frame, the closest paths to each vehicle ( Alg. \ref{alg:kdtree_assoc}) are recorded as a time-ordered association history. These records are used for predicting the future location and direction of each vehicle. 

\begin{algorithm}
\caption{\small{Future Path Estimation in Path-Index Space}}
\label{alg:future_path}
\fontsize{7.5}{7.7}\selectfont
\KwIn{$\mathcal{H}_v$, $\{\mathcal{R}_p\}$, $L,K,N$}
\KwOut{Predicted paths $\widehat{\mathcal{Y}}_v$}

$\widetilde{\mathcal{H}}_v\gets\text{Downsample}(\mathcal{H}_v;L,K)$\;
$\mathcal{P}\gets\{p\mid(p,i)\in\widetilde{\mathcal{A}}_v^{(1)}\}$\;
\ForEach{$\widetilde{\mathcal{A}}_v^{(k)}\in\widetilde{\mathcal{H}}_v$}{
    $\mathcal{P}\gets\mathcal{P}\cap\{p\mid(p,i)\in\widetilde{\mathcal{A}}_v^{(k)}\}$\;
}

$\widehat{\mathcal{Y}}_v\gets\emptyset$\;
\ForEach{$p\in\mathcal{P}$}{
    $\Delta i_p\gets\{i_p^{(k-1)}-i_p^{(k)}\},\;
    v_p\gets\mathrm{mean}(\Delta i_p),\;
    i_p^{\mathrm{last}}\gets i_p^{(1)}$\;
    \For{$n=1$ \KwTo $N$}{
        $\hat{i}_p^{(n)}\gets i_p^{\mathrm{last}}+n v_p$\;
        $\hat{\mathbf{r}}_p^{(n)}\gets\mathcal{R}_p[\hat{i}_p^{(n)}]\;\text{or extrapolate}$\;
    }
    $\widehat{\mathcal{Y}}_v\cup\{(p,\hat{\mathbf{r}}_p^{(1:N)})\}$\;
}
\Return $\widehat{\mathcal{Y}}_v$
\end{algorithm}

Alg.~\ref{alg:future_path} predicts vehicle motion in a path-index domain using historical path–index associations. Relying on a single association is unreliable, as vehicles may temporarily approach neighboring paths at junctions or merges without following them, leading to incorrect assignments. To address this, the algorithm applies temporal downsampling to the association history, producing $\widetilde{\mathcal{H}}_v$, which preserves multiple observations while reducing noise. A consistent path set $\mathcal{P}$ is then obtained by intersecting path labels across all retained history elements, ensuring that only persistently associated paths are considered.

For each $p \in \mathcal{P}$, motion is modeled in index space rather than Euclidean space. Index differences $\Delta i_p$ are computed across successive history elements to estimate an average index-space velocity $v_p$, and future indices are extrapolated from the most recent index $i_p^{\mathrm{last}}$ and mapped back to spatial coordinates using the path definition $\mathcal{R}_p$. Unlike single-step extrapolation, which is sensitive to noise, missed detections, and transient misassociations, the proposed multi-sample strategy enforces temporal consistency and yields more robust future path predictions, particularly in complex road geometries with closely spaced but topologically distinct paths.

\subsection{Collision Prediction Probability}

In the final step, probability future collision is computed by analyzing the spatiotemporal proximity of the predicted future vehicle trajectories. The algorithm is presented in Alg. \ref{alg:collision_prob}.

\begin{algorithm}
\caption{\small{Collision Probability Estimation}}
\label{alg:collision_prob}
\fontsize{7.5}{7.7}\selectfont
\KwIn{$\widehat{\mathcal{Y}}=\{v\mapsto(p\mapsto \widehat{\mathbf{R}}_{v,p})\}$, $D$, $\Delta t$, $T$}
\KwOut{Collision summary $\mathcal{S}$ with $\Pr_{\mathrm{col}}(v_1,v_2)$}

$\widehat{\mathcal{Y}} \gets \{\, v\mapsto(p\mapsto \widehat{\mathbf{R}}_{v,p}) \in \widehat{\mathcal{Y}} \;|\; |\widehat{\mathbf{R}}_{v,p}|>0 \,\}$\;
$\mathcal{S}\gets\emptyset,\;\mathcal{V}\gets\{v:\widehat{\mathcal{Y}}(v)\neq\emptyset\}$\;

\ForEach{$(v_1,v_2)\in\mathrm{Comb}(\mathcal{V},2)$}{
    $\mathcal{P}_1\gets\{p:|\widehat{\mathbf{R}}_{v_1,p}|>0\},\;
     \mathcal{P}_2\gets\{p:|\widehat{\mathbf{R}}_{v_2,p}|>0\}$\;
    $N_{\mathrm{comb}}\gets|\mathcal{P}_1|\cdot|\mathcal{P}_2|,\;
     N_{\mathrm{col}}\gets 0,\;
     \mathrm{ex}\gets\varnothing$\;

    \ForEach{$p_1\in\mathcal{P}_1$}{
        $M_1\gets|\widehat{\mathbf{R}}_{v_1,p_1}|$\;
        \If{$M_1<2$}{\textbf{continue}}
        $\widehat{\Gamma}_1\gets\{(t_1^{(k)},\mathbf{x}_1^{(k)})\}_{k=1}^{M_1},\;
        t_1^{(k)}=\frac{k-1}{M_1-1}T$\;
        
        $\mathbf{x}_1^{(k)}=\widehat{\mathbf{R}}_{v_1,p_1}[k]$\;

        \ForEach{$p_2\in\mathcal{P}_2$}{
            $M_2\gets|\widehat{\mathbf{R}}_{v_2,p_2}|$\;
            \If{$M_2<2$}{\textbf{continue}}
            $\widehat{\Gamma}_2\gets\{(t_2^{(\ell)},\mathbf{x}_2^{(\ell)})\}_{\ell=1}^{M_2},\;
            t_2^{(\ell)}=\frac{\ell-1}{M_2-1}T$\;
            
            $\mathbf{x}_2^{(\ell)}=\widehat{\mathbf{R}}_{v_2,p_2}[\ell]$\;

            $\mathcal{K}_2 \gets \mathrm{KDTree}(\{\mathbf{x}_2^{(\ell)}\}_{\ell=1}^{M_2})$, $c\gets 0$\;

            \ForEach{$(t_1,\mathbf{x}_1)\in\widehat{\Gamma}_1$}{
                $\mathcal{I}\gets\mathcal{K}_2.\mathrm{query\_point}(\mathbf{x}_1,D)$\;
                \ForEach{$\ell\in\mathcal{I}$}{
                    $(t_2,\mathbf{x}_2)\gets\widehat{\Gamma}_2[\ell]$\;
                    \If{$|t_1-t_2|\le\Delta t$}{
                        $c\gets 1$\;
                        \If{$\mathrm{ex}=\varnothing$}{
                            $\bar{t}\gets\frac{t_1+t_2}{2}$\;
                            
                            $\bar{\mathbf{x}}\gets \mathrm{round}\big((\mathbf{x}_1+\mathbf{x}_2)/2\big)$\;
                            $\mathrm{ex}\gets(p_1,p_2,\bar{t},\bar{\mathbf{x}})$\;
                        }
                        \textbf{break}\;
                    }
                }
                \If{$c=1$}{\textbf{break}}
            }

            $N_{\mathrm{col}}\gets N_{\mathrm{col}}+c$\;
        }
    }

    \If{$N_{\mathrm{col}}>0$}{
        $\Pr_{\mathrm{col}}(v_1,v_2)\gets \frac{N_{\mathrm{col}}}{N_{\mathrm{comb}}}$\;
        $\mathcal{S}\gets \mathcal{S}\cup\{(v_1,v_2,\Pr_{\mathrm{col}}(v_1,v_2),N_{\mathrm{comb}},N_{\mathrm{col}},\mathrm{ex})\}$\;
    }
}
\Return $\mathcal{S}$
\end{algorithm}

Alg. \ref{alg:collision_prob} estimates the probability of collisions by jointly considering the spatial and temporal alignment of predicted vehicle trajectories. Relying on spatial proximity alone is insufficient for collision assessment, as two vehicles may traverse the same location at different times due to the different starting point or speed. In such cases, although their future paths intersect geometrically, no collision occurs because one vehicle may have already passed the intersection point before the other arrives. Therefore, both spatial closeness and temporal coincidence must be satisfied to infer a collision risk.

After filtering invalid predicted trajectories from $\widehat{\mathcal{Y}}$, such as yield empty or degenerate predicted sequences, the algorithm evaluates all unordered vehicle pairs $(v_1,v_2)$ and their corresponding candidate future paths. Each predicted path is parameterized over a continuous time horizon $T$, producing time-stamped trajectories that encode not only where a vehicle will be, but also when it will be there. For each path pair $(p_1,p_2)$, a K-D tree enables efficient detection of spatially proximate trajectory points within distance threshold $D$ in lower inference time.
A potential collision is declared only when spatial proximity is accompanied by temporal alignment, enforced through the constraint $|t_1 - t_2| \le \Delta t$. This temporal tolerance accounts for variations in vehicle speed and prediction uncertainty, while preventing false positives caused by asynchronous path crossings. By combining spatial and temporal constraints, the algorithm ensures that detected collisions correspond to physically plausible interactions rather than mere geometric overlaps.
The final collision probability is computed as the ratio of colliding path combinations to the total number of evaluated combinations. This spatiotemporal formulation yields a robust and realistic measure of collision risk that faithfully reflects both vehicle motion dynamics and timing differences among predicted trajectories.

\section{Results and Discussion}

\begin{figure*}
    \centering
    \includegraphics[width=0.9\linewidth]{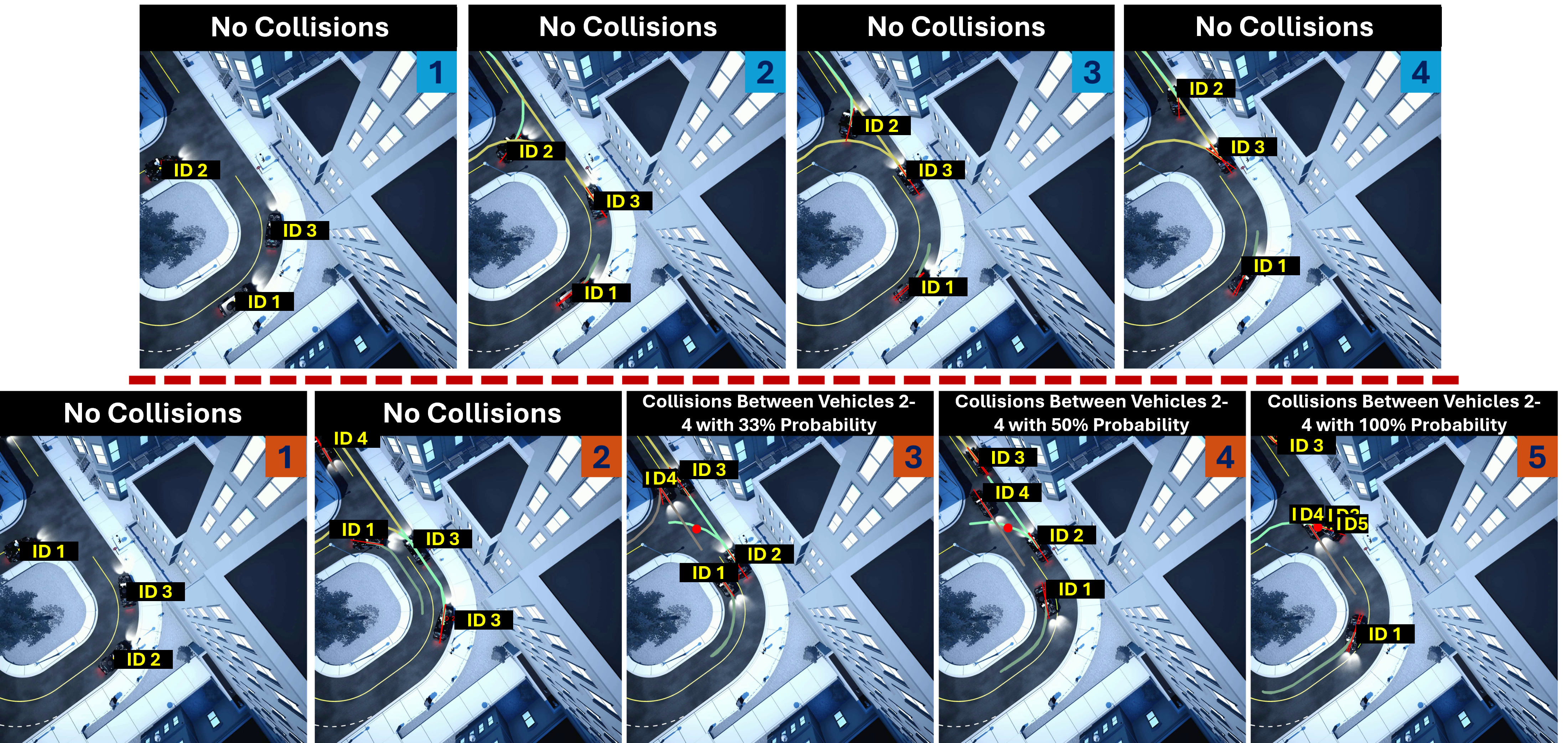}\vspace{0.1cm}
    \caption{Results on Vehicle Tracking and Collision Prediction in Two Scenarios: Safe Driving (Top) and Collision (Bottom)}
    \label{fig:environment1}
\end{figure*}

For evaluation, data are collected using the QLabs urban traffic digital twin (see Section~\ref{sec: quanser}). A total of 100 video sequences are generated at a resolution of $2048 \times 2048$ pixels, each with a maximum duration of 15 seconds. Fig.~\ref{fig:environment1} presents runtime results for safe driving and collision scenarios.

In Fig.~\ref{fig:environment1}, the top row shows a normal driving scenario, while the bottom row presents a collision sequence across consecutive frames. Red straight lines indicate vehicle's length and motion direction along the road, while colored trajectories represent possible future paths inferred by the proposed framework. In both scenarios, vehicles are assigned consistent identifiers throughout runtime.
In the safe driving example, the second frame shows vehicle ID~3 with two plausible future directions: continuing straight or turning left. Although predicted future paths of vehicle ID~3 spatially intersect with those of vehicle ID~2, the collision probability remains $0\%$ due to misalignment in their temporal indices, indicating that the vehicles reach the intersection point at different times. In contrast, in the collision scenario, vehicles with IDs~2 and~4 exhibit a collision probability of $33\%$ at frame~3, where ID~4 proceeds straight and ID~2 turns left, causing both spatial and temporal thresholds to be exceeded. At frame~4, the direction of vehicle ID~4 becomes unambiguous, increasing the collision probability to $50\%$. In the final frame, both vehicles have a single feasible future direction, resulting in a $100\%$ collision probability.

Performance is evaluated by identifying collision scenarios in which the framework fails to predict an impending event. Across all experiments, the proposed method achieves an overall collision detection accuracy of approximately $88\%$. Relying solely on YOLO-based object detection enables deployment on edge devices with low inference latency and without cloud-based video transmission. However, this design also introduces sensitivity to detection errors and adversarial conditions affecting YOLO performance. For example, duplicate detections may occur after a collision, as observed in frame~5 of the collision scenario where a single vehicle is assigned IDs~2 and~5. Moreover, speed estimation based on pixel-wise displacement can fluctuate during sharp turns, even when actual vehicle speed remains constant. Finally, collision prediction depends on sufficient historical observations; when vehicles first appear in close proximity without prior context, collisions may not be anticipated. Despite these limitations, the framework demonstrates robust performance across diverse scenarios and serves as a strong, reproducible methodology for future research.

\section{Conclusions} \label{sec:6}
This work introduced a lightweight digital-twin-based framework for vehicle tracking and spatiotemporal collision prediction, providing a clear reference point for systematic comparison with future approaches. The framework constructs pixel-wise path representations from YOLO detections, associates vehicles with road segments, estimates motion from frame-level variations, and predicts future locations to compute collision probability. Experimental results show that the proposed framework predicts approximately 88\% of collision events before occurrence in controlled scenarios, while highlighting expected limitations under adversarial conditions and complex dynamics. Future work will build upon this baseline to improve robustness and resilience in more realistic traffic environments.

\section*{Acknowledgment }\label{Section6}
This work is supported in part by the MITACS Accelerate Program, the NSERC CREATE TRAVERSAL program, and the Ontario Research Fund–Research Excellence program (Grant ORF-RE012-026). The authors thank Quanser for supporting traffic data generation via QLabs.


\bibliographystyle{IEEEtran}


\end{document}